\documentclass{article}
\usepackage{ijcai21}

\usepackage{times}
\usepackage{soul}
\usepackage{url}
\usepackage[hidelinks]{hyperref}
\usepackage[utf8]{inputenc}
\usepackage[small]{caption}
\usepackage{graphicx}
\usepackage{amsmath}
\usepackage{amsthm}
\usepackage{booktabs}
\usepackage{algorithm}
\usepackage{algorithmic}
\usepackage{amssymb}
\usepackage{hyperref}      
\usepackage{amsfonts}      
\usepackage{nicefrac}      
\usepackage{microtype}     
\usepackage{balance}
\usepackage{color}

\newtheorem{theorem}{Theorem}

\newtheorem{lemma}{Lemma}
\newtheorem{lemmaA}{Lemma}[section]
\newtheorem{proposition}{Proposition}
\newtheorem{remark}{Remark}
\newtheorem{corollary}{Corollary}

\title{Average-Reward Reinforcement Learning with Trust Region Methods}

\author{
Xiaoteng Ma$^1$ \and
Xiaohang Tang$^2$ \and
Li Xia$^3$\footnote{Corresponding author: Li Xia, Jun Yang} \and 
Jun Yang$^1$\footnotemark[1] \And 
Qianchuan Zhao$^1$ \\
\affiliations
$^1$Department of Automation, Tsinghua University\\
$^2$Department of Statistical Science, University College London\\
$^3$Business School, Sun Yat-sen University
\emails
\ ma-xt17@mails.tsinghua.edu.cn, xiaohang.tang.20@ucl.ac.uk, xiali5@mail.sysu.edu.cn, yangjun603@mail.tsinghua.edu.cn, zhaoqc@tsinghua.edu.cn, 
}

\begin{document}

\maketitle

\begin{abstract}
Most of reinforcement learning algorithms optimize the discounted criterion which is beneficial to accelerate the convergence and reduce the variance of estimates. Although the discounted criterion is appropriate for certain tasks such as financial related problems, many engineering problems treat future rewards equally and prefer a long-run average criterion. In this paper, we study the reinforcement learning problem with the long-run average criterion. Firstly, we develop a unified trust region theory with discounted and average criteria and derive a novel performance bound within the trust region with the Perturbation Analysis (PA) theory. Secondly, we propose a practical algorithm named Average Policy Optimization (APO), which improves the value estimation with a novel technique named Average Value Constraint. Finally, experiments are conducted in the continuous control environment MuJoCo. In most tasks, APO performs better than the discounted PPO, which demonstrates the effectiveness of our approach. Our work provides a unified framework of the trust region approach including both the discounted and average criteria, which may complement the framework of reinforcement learning beyond the discounted objectives.
\end{abstract}

\section{Introduction}

The deep reinforcement learning (DRL) achieves prominent progress in many fields~\cite{silver2016mastering,vinyals2019grandmaster,senior2020improved}. Most of the modern RL algorithms
aim at maximizing $\mathbb{E} \left[\sum_t^\infty \gamma^t r_t \right]$, where $\gamma$ denotes a \emph{discount factor} strictly less than one. The discount factor undermines the contribution of future rewards on the current value and ignores the long-run gains~\cite{sutton2018reinforcement},
which means that the reward received after $k$ steps in the future is only worth $\gamma^{k-1}$ of it to the current state. The existence of discount factor simplifies the theoretical analysis by constructing the contraction mapping and offers a faster and more stable online learning by reducing estimation variances~\cite{kakade2001optimizing,marbach2003approximate}.
However, the discounted criterion is not consistent with the natural metric in many real scenarios, and introduces unnecessary bias of estimates despite lower variance~\cite{thomas2014bias,gae}. 
In empirical implementations, the discount factor serves as a hyper-parameter and the algorithm performance is very sensitive to it~\cite{duan2016benchmarking}. Thus, the decision maker should make a compromise between the foresight of the decision and the learning stability. 

Different from the discounted criterion, the average criterion directly focuses on the long-run average performance. In the everlasting real-world problems, such as manufacturing, power systems and traffic controls, the average criterion is best suited~\cite{mahadevan1996average,dewanto2020average}. Optimizing the average criterion completely removes the discrepancy between the training performance and the evaluating metric due to discounting. Therefore, the average criterion is well explored in the classical MDP literature~\cite{howard1960dynamic,puterman1994markov,cxr}.
However, for the data-driven implementation, the average-reward RL algorithms is much less investigated compared with the discounted criterion. It is relatively hard to study the average RL algorithms, since many properties are lost when setting $\gamma=1$ in the discounted framework. An obvious evidence is that the Bellman operator is no longer a contraction mapping in the average setting~\cite{bertsekas1995neuro}.

One of the key difficulties in the average-reward RL is to guarantee the values estimated from the sample paths bounded without discounting. Many previous works extended the standard Q-learning to the average setting, such as R-learning~\cite{r-learning},
RVI-learning~\cite{puterman1994markov,bertsekas1995dynamic}, 
CSV-learning~\cite{csv-learning}, and Differential Q-learning~\cite{wan2021learning}. In the methods above, a relative value (the average reward estimate, e.g., the reference state value or a constant from prior knowledge) is subtracted from the target value, while they may still cause divergence and lead to unstable training~\cite{dewanto2020average}. 
Compared with the value-iteration based ones, the policy-iteration based methods, such as policy gradient~\cite{sutton2000policy} and Actor-Critic~\cite{konda2000actor}, are perhaps more suitable for average reward algorithms due to more stable training and potential for large scale problems. However, they still suffer the instability problem of value estimation in the average setting. 

In this paper, we solve the average-reward RL problem by introducing the trust region approach for the average criterion.
Based on the \emph{Perturbation Analysis} (PA) theory~\cite{cao1997perturbation,cxr}, we overcome the analysis difficulty and develop a unified trust region theory with discounted and average criteria. 
Motivated by using the \emph{Kemeny's constant} to bound the trust region policy improvement in the average reward case \cite{zhang2021policy}, we provide a unified performance bound which tightens the previous one~\cite{trpo,cpo} when $\gamma \to 1$. The bound also provides a direct evidence that average criterion is preferred than the discounted one for ergodic MDPs. From the view of implementation, we develop a practical algorithm called Average Policy Optimization (APO). We address the difficulty of estimating value with the average criterion and provide a new method named Average Value Constraint to solve the value drifting problem. Our APO algorithm provides a novel implementation for the average DRL algorithm with neutral networks as approximators. Finally, our experiments on the continuous control benchmark MuJoCo show that APO can beat the discounted PPO in the metric of average rewards, which further confirms the superiority of our approach.

\section{Preliminaries}

\subsection{Average Reward MDPs}
Consider an infinite-horizon ergodic Markov decision process (MDP), defined by the tuple $\langle \mathcal{S}, \mathcal{A}, P, r, d_0 \rangle$, where $\mathcal{S}$ denotes a finite set of states, $\mathcal{A}$ denotes a finite set of actions, $P:\mathcal{S} \times \mathcal{A}\times\mathcal{S} \to [0, 1]$ denotes the transition probability, $r:\mathcal{S} \times \mathcal{A} \to \mathbb{R}$ denotes the reward function, $d_0:\mathcal{S} \to [0, 1]$ denotes the distribution of the initial state $s_0$. Let 
$\pi:\mathcal{S} \times \mathcal{A} \to [0,1]$ denote a stochastic policy, $\tau$ denote a trajectory $(s_0, a_0, s_1,\cdots)$, and $\tau \sim \pi$ denote that the distribution of trajectories depends on $\pi$: $s_0 \sim d_0, a_t \sim \pi(\cdot \mid s_t), s_{t+1} \sim P(\cdot \mid s_t, a_t)$.

With the assumption of ergodicity of MDPs, we aim at maximizing the long-run average performance under $\pi$:
\begin{equation}
\eta_\pi := \lim_{T \to \infty} \frac{1}{T} \mathbb{E}_{\tau \sim \pi} \left[ \sum_{t=0}^{T-1} r(s_{t}, a_{t}) \right].
\end{equation}
We are interested in the \textit{steady-state distribution} $d_\pi$, which is defined as follows:
\begin{equation}
d_{\pi}(s) := \lim_{T \to \infty} \frac{1}{T} \mathbb{E}_{\tau \sim \pi} \left[ \sum_{t=0}^{T-1} P(s_t=s)  \right].
\end{equation}
Note that since we focus on the contribution of each action within the infinite horizon, the long-run performance is independent of the initial state distribution. Let $V_{\pi}(s)$ denote the \textit{state-value function}, $Q_{\pi}(s,a)$ denote the \textit{action-value function}, and $A_{\pi}(s,a)$ denote the \textit{advantage function}. The value functions are defined by subtracting the long-run average reward $\eta_\pi$ from the current reward to make the functions bounded~\cite{howard1960dynamic}:
\begin{align}
V_{\pi}(s) &:= \mathbb{E}_{\tau \sim \pi} \left[\sum_{t=0}^{\infty} (r(s_{t}, a_{t}) - \eta_\pi) \mid s_0=s \right], \notag \\
Q_{\pi}(s, a) &:= \mathbb{E}_{\tau \sim \pi} \left[\sum_{t=0}^{\infty} (r(s_{t}, a_{t}) - \eta_\pi) \mid s_0=s, a_0=a \right], \notag \\
A_{\pi}(s, a) &:= Q_{\pi}(s, a)-V_{\pi}(s).
\end{align}

\subsection{Discounted Reward MDPs}
With the discounted reward criterion, we focus on the recent rewards by introducing the discount factor $\gamma \in (0,1)$. We optimize the policy to maximize its normalized expected discounted return:
\begin{equation}
\eta_{\pi, \gamma} := (1 - \gamma) \mathbb{E}_{\tau \sim \pi} \left[ \sum_{t=0}^{\infty} \gamma^{t} r(s_t, a_t)  \mid s_0 \sim d_0 \right].
\end{equation}
The discounted performance $\eta_{\pi, \gamma}$ can be viewed as the average reward under \textit{discounted steady-state distribution} $d_{\pi, \gamma}$:
\begin{equation}
d_{\pi, \gamma}(s) := (1- \gamma) \mathbb{E}_{\tau \sim \pi} \left[ \sum_{t=0}^\infty \gamma^t P(s_t=s) \mid s_0 \sim d_0 \right].
\end{equation}

Different from the average criterion, the policy performance with the discounted criterion is dependent on $d_0$. The discount factor makes a smooth interpolation between $d_0$ and $d_\pi$. To see that, we have $\lim_{\gamma \to 0} d_{\pi, \gamma} = d_0$ (the bandit case) and $\lim_{\gamma \to 1} d_{\pi, \gamma} = d_\pi$ (the long-run average reward case).

In the modern DRL literature, the value functions are defined as following:
\begin{align}
\label{equ:dis_val_old}
V_{\pi, \gamma}(s) &:= \mathbb{E}_{\tau \sim \pi} \left[ \sum_{t=0}^{\infty} \gamma^t r(s_{t}, a_{t}) \mid s_0=s \right], \notag \\
Q_{\pi, \gamma}(s, a) &:= \mathbb{E}_{\tau \sim \pi} \left[ \sum_{t=0}^{\infty} \gamma^t r(s_{t}, a_{t}) \mid s_0=s, a_0=a \right], \notag  \\
A_{\pi, \gamma}(s, a) &:= Q_{\pi, \gamma}(s, a) - V_{\pi, \gamma}(s).
\end{align}
As the value function is a core concept in RL, people may expect to optimize the long-run performance by letting $\gamma \to 1$. Unfortunately, many important properties are lost when $\gamma \to 1$, e.g., value functions will be unbounded which devastate the algorithm performance. Another example is the trust region theory for policy optimization, which we will discuss next.

\subsection{Trust Region Method}
The trust region methods~\cite{trpo,ppo,cpo} are a series of DRL algorithms updating policies with the following approximate policy iteration:
\begin{align}
    \label{equ:problem}
    \pi_{k+1} = & \mathop{\arg\max}_{\pi} \mathbb{E}_{s \sim d_{\pi_k, \gamma}, a \sim \pi} \left[ A_{\pi_k, \gamma}(s,a) \right] \\
    & \text{s.t. } \bar{D}(\pi \parallel \pi_k) \leq \delta, \notag
\end{align}
where $\bar{D}(\pi \parallel \pi_k) = \mathbb{E}_{s \sim d_{\pi_k, \gamma}} [D(\pi \parallel \pi_k)[s]]$ with $D$ denoting some distance measure and $\delta > 0$ is the step size. The policy set $\left\{\pi \mid \bar{D}(\pi \parallel \pi_k) \leq \delta \right\}$ is called \textit{trust region}, in which it is safe to evaluate policies with the samples collected with $\pi_k$. 
In theoretical analysis, $D $ is often chosen as the \textit{total variation divergence} defined by $D_{\text{TV}}(\pi^\prime \parallel \pi)[s] = \frac{1}{2} \sum_{a \in \mathcal{A}} | \pi^\prime( a \mid s)- \pi(a \mid s) | $. In practice, \textit{Kullback-Leibler divergence} is preferred, which is defined as $D_{\text{KL}}(\pi^\prime \parallel \pi)[s] = \sum_{a \in \mathcal{A}} \pi^\prime( a \mid s) \log \frac{\pi^\prime( a \mid s)}{\pi(a \mid s)}$. Since $D_{\text{TV}}(p \parallel q) \leq \sqrt{D_{\text{KL}}(p \parallel q) / 2}$, the theoretical analysis is aligned with the practical implementation.

The primary motivation of the trust region method is that optimizing the surrogate objective in the trust region guarantees monotonic performance improvements. When the distance is bounded by $D_{\text{TV}}$, the solution of the problem in (\ref{equ:problem}) has the lower performance bound~\cite{cpo}
\begin{align*}
    & \eta_{\pi_{k+1}, \gamma} - \eta_{\pi_{k}, \gamma} \geq  \\
    & \mathbb{E}_{s \sim d_{\pi_{k+1}, \gamma}, a \sim \pi_{k+1}} \left[ A_{\pi_k, \gamma}(s,a)- \frac{2 \gamma \epsilon_\gamma}{1 - \gamma} D_{\text{TV}} (\pi_{k+1} \parallel \pi_k)[s] \right],
\end{align*}
where $\epsilon_\gamma = \max_s | \mathbb{E}_{a \sim \pi_{k+1}} [A_{\pi_k, \gamma}(s,a)] |$.

The advantage of the trust region method majorly comes from two aspects. In each iteration, the samples collected with $\pi_k$ are reused in evaluating any $\pi$ in trust region, which significantly improves the data efficiency compared with the traditional policy gradient methods. 
The other hand, the performance lower bound avoids collapse after bad updates when neural networks are used as the policy approximator~\cite{duan2016benchmarking}.

However, existing analysis of the lower performance bound for the trust region method is not applicable when $\gamma \to 1$, which can be verified by $\lim_{\gamma \to 1} \gamma / (1-\gamma) = \infty$. As the larger discounted factor and average reward setting are important in real problems, more delicate theoretical analysis is needed to understand the behavior of trust region method when $\gamma \to 1$ even $\gamma = 1$.

\section{Unified Trust Region Theory with Discounted and Average Criteria}
In this section, we propose a unified trust region theory based on the PA theory that extends the existing analysis of performance improvement bound for both the average and discounted criteria. Moreover, we provide a novel policy performance bound for police updating. With the help of this bound, performance monotonicity of policy improvement is guaranteed when $\gamma \to 1$.
\subsection{General Formulation}

We redefine the discounted value functions in (\ref{equ:dis_val_old}) as follows:
\begin{small}
\begin{align}
\label{equ:dis_val_new}
V_{\pi, \gamma}(s) &:= \mathbb{E}_{\tau \sim \pi} \left[ \sum_{t=0}^{\infty} \gamma^t \left(r(s_{t}, a_{t}) - \eta_\pi \right) \mid s_0=s \right], \notag \\
Q_{\pi, \gamma}(s, a) &:= \mathbb{E}_{\tau \sim \pi} \left[ \sum_{t=0}^{\infty} \gamma^t \left(r(s_{t}, a_{t}) - \eta_\pi \right) \mid s_0=s, a_0=a \right], \notag  \\
A_{\pi, \gamma}(s, a) &:= Q_{\pi, \gamma}(s, a) - V_{\pi, \gamma}(s).
\end{align}
\end{small}
Although this formulaic approach may be not familiar to DRL researchers, it is commonly used in the MDP literature~\cite{cxr}. The main benefit of this value definition is that it bridges the gap between the discounted and average criteria. With the new definition of value functions, we obtain that $\lim_{\gamma \to 1} V_{\pi, \gamma} = V_\pi$ immediately.

Next, we present the \textit{performance difference formula} for any policies $\pi, \pi^\prime$.
\begin{lemma}
For any policies $\pi, \pi^\prime$, the following equation holds:
\begin{equation}
\label{equ:discounted-identity}
\eta_{\pi^\prime, \gamma} - \eta_{\pi, \gamma} = \mathbb{E}_{s \sim d_{\pi^\prime, \gamma}, a \sim \pi^\prime} \left[A_{\pi,\gamma}(s,a) \right].
\end{equation}
\end{lemma}
This relationship has been shown in many previous studies of trust region method~\cite{kakade2002approximately,trpo,cpo}. Actually, (\ref{equ:discounted-identity}) can be viewed as a rewriting of the performance difference formula in a sample path form. The performance difference formula is the key result of PA theory~\cite{cxr} and it quantifies the change of the performance of Markov systems corresponding to the change of policies. For the readers unfamiliar with PA, we attach a brief introduction in Appendix~A.1.

Powered by the performance difference formula, we derive the average performance difference formula at once.
\begin{corollary}
For any policies $\pi, \pi^\prime$, the following equation holds:
\begin{equation}
\label{equ:average-identity}
\eta_{\pi^\prime} - \eta_\pi = \mathbb{E}_{s \sim d_{\pi^\prime}, a \sim \pi^\prime} \left[A_\pi(s,a) \right].
\end{equation}
\end{corollary}

\subsection{Policy Improvement Bound}
While the performance difference formula accurately describes the performance difference for two arbitrary policies, it is difficult to use the formula to develop practical algorithms directly. The bottleneck is that to evaluate the performance difference, we need the samples from $d_{\pi^\prime, \gamma}$. That is self-contradictory since no trajectory of the new policy is available unless the policy is updated.
Thus, instead of optimizing the difference formula directly, the trust region method optimizes the following surrogate objective:
\begin{equation}
    L_{\pi, \gamma}(\pi^\prime) := \mathbb{E}_{s \sim d_{\pi, \gamma}, a \sim \pi} \left[\frac{\pi^\prime(a \mid s)}{\pi(a \mid s)} A_{\pi, \gamma}(s,a) \right],
\end{equation}
where $d_{\pi^\prime, \gamma}$ is replaced by $d_{\pi, \gamma}$ in (\ref{equ:discounted-identity}) and the probability of the sampled actions is corrected by importance sampling.

It is natural to query what is the difference between the surrogate objective and the real performance difference. To answer this question, we present the bound as follows.
\begin{proposition}
    For any two stochastic policies $\pi, \pi^\prime$, the following bound holds:
    \begin{align}
        \eta_{\pi^\prime, \gamma} - \eta_{\pi, \gamma} \geq L_{\pi, \gamma}(\pi^\prime) -  2 \epsilon_\gamma D_{\text{TV}}(d_{\pi^\prime, \gamma} \parallel d_{\pi, \gamma}), \notag \\
        \eta_{\pi^\prime, \gamma} - \eta_{\pi, \gamma} \leq L_{\pi, \gamma}(\pi^\prime) +  2 \epsilon_\gamma D_{\text{TV}}(d_{\pi^\prime, \gamma} \parallel d_{\pi, \gamma}),
    \end{align}
where $\epsilon_\gamma = \max_s | \mathbb{E}_{a \sim \pi^\prime} [A_{\pi, \gamma}(s,a)] |$.
\end{proposition}
The result tells us that if the discounted steady-state distribution of two policies is close enough and the advantage function is bounded, we are safe to use the surrogate objective for policy updating. 

Next, we build up the connection between the distance of the discounted steady-state distributions and the distance of policies.

\begin{proposition}
    For any two stochastic policies $\pi, \pi^\prime$, the following bound holds:
    \begin{equation}
        D_{\text{TV}}(d_{\pi^\prime, \gamma} \parallel d_{\pi, \gamma}) \leq \xi_\gamma \mathbb{E}_{s \sim d_{\pi, \gamma} } [D_{\text{TV}}(\pi^\prime \parallel \pi)[s]],
    \end{equation}
where $\xi_\gamma = \min \left\{ \frac{\gamma}{1 - \gamma}, \left| \frac{\gamma(\kappa_{\pi^\prime} -1)}{1 - (1-\gamma)\kappa_{\pi^\prime}} \right| \right\} $ and $\kappa_{\pi^\prime}$ is the Kemeny’s constant of the Markov chain induced by $\pi^\prime$.
\end{proposition}

The above conclusion is the key contribution of this paper, which reveals that \emph{the difference of discounted steady-state distribution is bounded with the distance of the policies multiplied with a constant factor $\xi_{\gamma}$}. This factor $\xi_{\gamma}$ is a time constant that reflects how policy changes affect long-term behaviors of Markov chains. When $\gamma$ is not very large, the long-run effect of the current decision gradually decreases, allowing decision makers to focus on a few steps after the decision and ignore the impact afterwards. However, if $\gamma$ is large enough, we cannot use discounts to suppress the long-term impact of the policy anymore. Fortunately, when $\gamma \to 1$, we find that it is still able to bound the long-run impact by introducing \textit{Kemeny's constant}~\cite{kemeny1960finite}, which means the average time of returning to the steady-state distribution starting from an arbitrary state. The idea of using Kemeny's constant to bound the distance of steady-state distributions is first introduced by Zhang and Ross~\cite{zhang2021policy}, and we extend it to a more general case with both the average and discounted criteria.

We further explore the theoretical result of $\xi_\gamma$ by observing following phenomena.
\begin{remark}
    For the average reward criterion, we have $\xi = \lim_{\gamma \to 1} \xi_\gamma = \kappa_{\pi^\prime} - 1$, which is consistent with~\cite{zhang2021policy}.
\end{remark}

\begin{remark}
    The $\xi_\gamma$ is bounded by a $\gamma$-independent constant:
    \begin{equation}
        \xi_\gamma \leq 2(\kappa_{\pi^\prime} - 1), \forall \gamma \in [0, 1].    
    \end{equation}
     In particular, the maximum is not taken at $\gamma =1$ but
    \begin{equation}
        \label{equ:max_xi}
        \mathop{\arg\max}_{\gamma} \xi_\gamma = 1 - \frac{1}{2 \kappa_{\pi^\prime} - 1}.
    \end{equation}
\end{remark}

Remark 2 says that we are always safe to implement the trust region approach whatever discount factor we have. We may not intuitively believe that a smaller $\gamma$ always leads to a tighter performance bound. When $\gamma$ is larger than (\ref{equ:max_xi}), the lower bound decreases as the $\gamma$ increases and finally becomes $\kappa_{\pi^\prime} - 1$ at $\gamma = 1$. It supports that when the problem focuses more on long-term performance, the average criterion is better than a large discount factor in the trust region method, since the former is able to obtain a more accurate estimation.

Combining the preceding two bounds together, we conclude the following theorem.
\begin{theorem}
    For any two stochastic policies $\pi, \pi^\prime$, the following bound holds:
    \begin{align*}
        \eta_{\pi^\prime, \gamma} - \eta_{\pi, \gamma} \geq L_{\pi, \gamma}(\pi^\prime) -  2 \epsilon_\gamma \xi_\gamma \mathbb{E}_{s \sim d_{\pi, \gamma} } [D_{\text{TV}}(\pi^\prime \parallel \pi)[s]], \\
        \eta_{\pi^\prime, \gamma} - \eta_{\pi, \gamma} \leq L_{\pi, \gamma}(\pi^\prime) +  2 \epsilon_\gamma \xi_\gamma \mathbb{E}_{s \sim d_{\pi, \gamma} } [D_{\text{TV}}(\pi^\prime \parallel \pi)[s]].
    \end{align*}
    In particular, the bounds hold with the average criterion:
    \begin{align*}
        \eta_{\pi^\prime} - \eta_\pi \geq L_\pi(\pi^\prime) - 2 \epsilon \xi \mathbb{E}_{s \sim d_\pi } [D_{\text{TV}}(\pi^\prime \parallel \pi)[s]], \\
        \eta_{\pi^\prime} - \eta_\pi \leq L_\pi(\pi^\prime) + 2 \epsilon \xi \mathbb{E}_{s \sim d_\pi } [D_{\text{TV}}(\pi^\prime \parallel \pi)[s]].
    \end{align*}
\end{theorem}
The result clearly shows that the performance difference is bounded by three components: \emph{the difference of policies, the maximum change in the values, and a time constant depending on the problem}. It not only facilitates the design of practical trust region algorithms, but also inspires us to further understand the nature of general RL algorithms.

In the end of this section, we give another perspective for why the average criterion is preferred to the discounted one in trust region method. It should be emphasized that the above analysis is based on the discounted steady-state distribution, from which it is not convenient to sample. In the practical implementation, most discounted algorithms ignore the difference in state distributions and directly sample from the steady-state distribution $d_\pi$, which leads to biased estimates for the policy gradients ~\cite{thomas2014bias}. However, with the average criterion, performance estimation is unbiased as sampling under $d_\pi$ is theoretically-justified.

\section{Average Policy Optimization}
Based on the unified trust region theory, we extend the current discounted trust region algorithms to the average reward criterion. We address the value drifting problem raised in the average setting, and present a technique named Average Value Constraint for better value estimation. By approximating the policy and value function with neutral networks, we develop an algorithm based on the Actor-Critic framework, which is named Average Policy Optimization.

\subsection{Value Estimation}

In the most DRL algorithms, the value function is approximated by a value network $V_\phi(s)$ with $\phi$ as the parameters, which is updated by minimizing the following loss with stochastic gradient descent (SGD) method
\begin{equation}
    \label{equ:ori_problem}
    \min_\phi J_V(\phi) =  \frac{1}{N} \sum_{n=1}^N \left[ \frac{1}{2} (\hat{V}(s_n) - V_\phi(s_n))^2 \right],
\end{equation}
where $n$ indexes a batch of states sampled with $\pi$ and $\hat{V}(s_n)$ is the target value for $s_n$.

Under the discounted criterion, $\hat{V}(s_n)$ can be calculated directly using Monte Carlo (MC) method with $\hat{V}_{\text{MC}}(s_n) = \sum_{t=0}^\infty \gamma^t r_{n+t}$, or using the value of next step to bootstrap (TD): $\hat{V}_\phi(s_n) = r_n + \gamma V_\phi(s_{n+1})$~\cite{sutton2018reinforcement}, where the latter is preferred in most cases for low-variance estimates. In the average setting, we adopt the update rule of R-learning~\cite{r-learning} to evaluate the target value with $\hat{V}_\phi(s_n) = r_n - \hat{\eta} + V_\phi(s_{n+1})$, where $\hat{\eta}$ is the estimate of $\eta_\pi$. To smooth the value of $\hat{\eta}$ over 
batches, we update $\hat{\eta}$ by a moving average method: $\hat{\eta} \gets (1 - \alpha) \hat{\eta} + \alpha \frac{1}{N} \sum_{n=1}^N r_n$, where $\alpha$ is the step size. 

Let $\phi^\prime$ denote the updated parameters:
\begin{equation}
    \phi^\prime = \phi + \beta (\hat{V}_\phi(s)- V_\phi(s)) \nabla V_\phi(s),
\end{equation}
where $\beta$ is the learning rate. As the value network is updated by SGD, noise in the estimate is inevitable. Suppose that there is a bias $\epsilon_V$ between the value function and the true value $V_\phi(s) = V_{\tilde{\phi}}(s) + \epsilon_V, \forall s \in \mathcal{S}$. The real updated parameters based on the $V_{\tilde{\phi}}$ is
\begin{equation}
    \tilde{\phi}^\prime = \phi + \beta (\hat{V}_{\tilde{\phi}}(s) - V_{\tilde{\phi}}(s)) \nabla V_{\tilde{\phi}}(s).
\end{equation}
As $\beta$ is sufficiently small, the post-update value function can be well-approximated by linearizing around $\phi$ using Taylor’s expansion:
\begin{align*}
    V_{\phi^\prime}(s) & \approx V_\phi(s) + \beta (\hat{V}_\phi(s) - V_\phi(s)) \|\nabla{V}_\phi(s)\|_2^2, \\
    V_{\tilde{\phi}^\prime}(s) & \approx V_{\tilde{\phi}}(s) + \beta (\hat{V}_{\tilde{\phi}}(s) - V_{\tilde{\phi}}(s)) \|\nabla{V}_{\tilde{\phi}}(s)\|_2^2. 
\end{align*}
Subtracting the first equation from the second one, we have
\begin{equation}
     V_{\phi^\prime}(s) -  V_{\tilde{\phi}^\prime}(s) \approx \left(1 -\beta (1-\gamma) \|\nabla_\phi{V}_\phi(s)\|_2^2 \right) \epsilon_V.
\end{equation}
For $\gamma < 1$, the bias between the real values and approximated values is gradually eliminated during training. However, this feedback mechanism disappears for $\gamma=1$ as $V_{\phi^\prime}(s) - V_{\tilde{\phi}^\prime}(s) \approx \epsilon_V$. That means the approximated values are drifting away from the true values during training due to noises. We call it the \emph{value drifting problem} in the average reward setting.

The analysis above explains why the value estimation in average setting is inaccurate in the data-driven mode. Next we try to correct the estimation by taking advantage of the following property.
\begin{proposition}
    $\mathbb{E}_{s \sim d_\pi} \left[V_{\pi, \gamma}(s) \right] = 0$.  
\end{proposition}
It tells us that the ideal estimates of values defined by (\ref{equ:dis_val_new}) should have zero mean. We rewrite the problem in (\ref{equ:ori_problem}) by adding this condition as a constraint:
\begin{align}
    \label{equ:modified_problem}
    &\min_\phi J_V(\phi) = \frac{1}{N} \sum_{n=1}^N \left[ \frac{1}{2} (\hat{V}(s_n) - V_\phi(s_n))^2 \right], \notag \\
    & \text{s.t. } \frac{1}{N} \sum_{n=1}^N \left[V_\phi(s_n) \right] = 0.
\end{align}
By constructing the Lagrangian function of~(\ref{equ:modified_problem}), we have
\begin{equation*}
    \min_\phi L_V(\phi, \nu) = \frac{1}{N} \sum_{n=1}^N \left[ \frac{1}{2} (\hat{V}(s_n) - V_\phi(s_n))^2 \right]- \frac{1}{2} \nu b^2.
\end{equation*}
where $b := \frac{1}{N} \sum_{n=1}^N \left[V_\phi(s_n) \right]$ and $\nu$ is the Lagrangian multiplier. Taking the gradient of $L_V(\phi, \nu)$, we obtain
\begin{small}
\begin{align*}
    &\nabla L_V(\phi, \nu) \\
    &= \frac{1}{N} \sum_{n=1}^N \left[ (\hat{V}(s_n) - V_\phi(s_n)) \nabla_\phi V_\phi(s_n) \right]-  \frac{\nu b}{N} \sum_{n=1}^N \nabla_\phi [ V_\phi(s_n) ] \\
    &= \frac{1}{N} \sum_{n=1}^N \left[ (\hat{V}(s_n) - \nu b - V_\phi(s_n)) \nabla_\phi V_\phi(s_n) \right].
\end{align*}
\end{small}
The problem in (\ref{equ:modified_problem}) is equivalent to the following problem 
\begin{equation} \label{equ:J_V}
    \min_\phi J_V(\phi) = \frac{1}{N} \sum_{n=1}^N \left[ \frac{1}{2} (\tilde{V}(s_n) - V_\phi(s_n))^2 \right],
\end{equation}
where $\tilde{V}(s_n) := \hat{V}(s_n) - \nu b$. Compared with the discounted variant, we explicitly control the bias in value estimation. We name this method as Average Value Constraint.

\subsection{Policy Optimization}
Many algorithms~\cite{trpo,ppo,wu2017scalable} have been proposed to approximately solve the discounted problem in (\ref{equ:problem}) with a policy network. Here we choose PPO~\cite{ppo} as the base to develop our average reward algorithm, and point out that the policy updating is similar with other trust region methods. Instead of optimizing the original objective with the constraint, we optimize the surrogate loss as follows:
\begin{small}
\begin{equation}\label{equ:J_pi}
    J_\pi(\theta) =
    \frac{1}{N }\sum_{n=1}^N \left[
    \min \left(\omega(\theta) \hat{A}_n, 
    \text{clip}( \omega(\theta),
    1-\varepsilon,
    1+\varepsilon)\hat{A}_n \right) \right],
\end{equation}
\end{small}
where $\theta$ is the parameters of the policy network, $\omega(\theta) = \frac{\pi_\theta(a_t \mid s_t)}{\pi(a_t \mid s_t)}$ is the importance sampling ratio and $\hat{A}_n$ is a shorthand of advantage estimation $\hat{A}(s_n, a_n)$. To better balance the variance and bias of policy gradients, we further modify the discounted GAE~\cite{gae} to an average variant. Define the average TD residual as
\begin{equation}\label{equ:td}
    \delta_t = r(s_t, a_t)-\hat{\eta} + V_\phi(s_{t+1}) - V_\phi(s_t).    
\end{equation}
Then the advantage estimator is
\begin{equation}\label{equ:adv}
    \hat{A}(s_n,a_n)= \sum_{t=0}^{\infty} \lambda^{t}\delta_{n+t},    
\end{equation}
where $\lambda$ provides the trade-off between bias and variance.

\begin{algorithm}[tb] 
    \caption{Average Policy Optimization}
    \label{alg2}
    \textbf{Input}: $\alpha, \beta, \lambda, \nu, \varepsilon, K, N, M$
    \begin{algorithmic}[1]
        \STATE Initialize $\theta, \phi$ randomly.
        \STATE Set $\hat{\eta} = 0$, $b = 0$.
        \FOR {$k=1,2,\cdots, K$} 
        \STATE Run policy $\pi_{\theta_k}$ for $N$ times to collect $\{s_n, a_n, r(s_n, a_n), s_{n+1}\}, n =1, \dots, N$.
        \STATE Update $\hat{\eta} \gets (1 - \alpha) \hat{\eta} + \alpha \frac{1}{N} \sum_{n=1}^N r(s_n, a_n)$.
        \STATE Update $ b \gets (1 - \alpha) b + \alpha \frac{1}{N} \sum_{n=1}^N V_\phi(s_n)$.
        \STATE Compute $\delta_n$ with (\ref{equ:td}) at all timesteps.
        \STATE Compute $\hat{A}(s_n,a_n)$ with (\ref{equ:adv}) at all timesteps.
        \STATE Update the $\theta$ with $J_\pi(\theta)$ in (\ref{equ:J_pi}) for $M$ epochs.
        \STATE Update the $\phi$ with $J_V(\phi)$ in (\ref{equ:J_V}) for $M$ epochs.
        \ENDFOR
    \end{algorithmic}
\end{algorithm}

\begin{figure*}[tbp]
    \centering
    \includegraphics[width=0.93\linewidth]{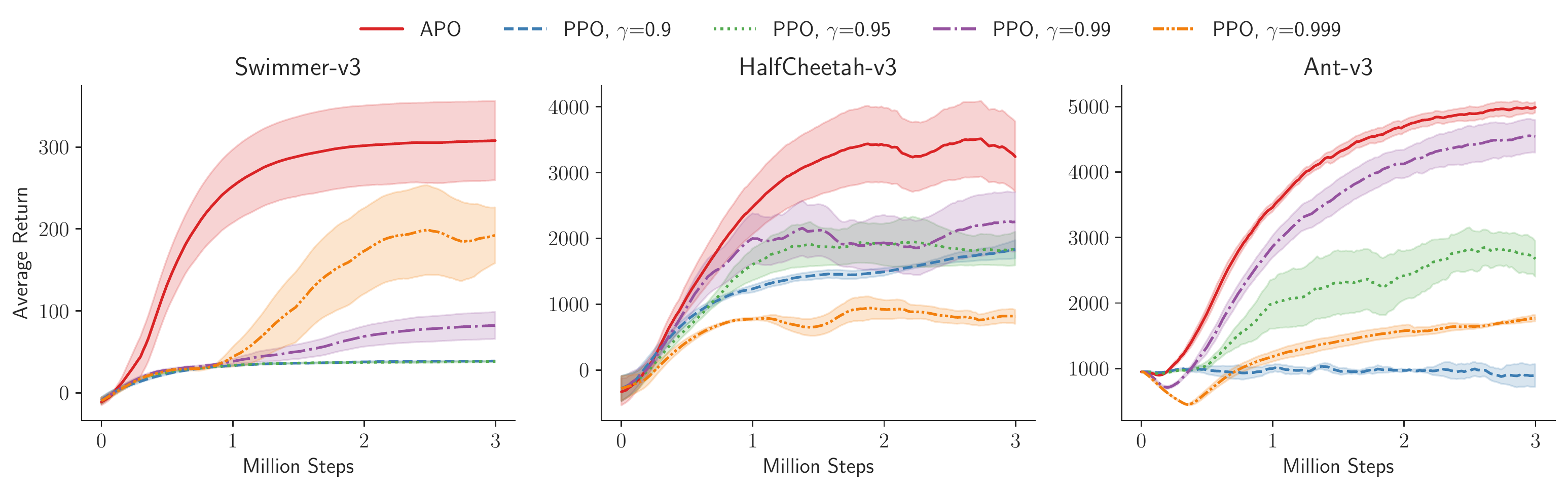}
    \vspace{-1em}
    \caption{Compare APO and PPO with different discount factors in MuJoCo tasks. Results for more tasks can be found in Appendix~C.}
    \label{fig:mujoco}
\end{figure*}

\section{Related Work}
Previous studies of average-reward RL mainly focus on $Q$-learning type algorithms in the tabular setting~\cite{r-learning,abounadi2001learning,csv-learning,wan2021learning}. For the policy gradient methods~\cite{baxter2001infinite,kakade2001optimizing}, they often use a discount factor approaching one to simulate the average case instead of optimizing the average objective directly. A rapid developing line of average-reward RL with function approximations 
~\cite{wei2021learning,zhang2021average,hao2021adaptive} contributes valuable insights under the linear function assumption. However, the average reward setting has not yet caught too much attention of the DRL community. \\\\
Concurrently, there are new advances in the trust region approach for average-reward RL~\cite{zhang2021policy,dai2021refined}. Zhang and Ross~\cite{zhang2021policy} first introduced the Kemeny's constant in the bound of average reward trust region RL, which provided an interesting and important way to bound the performance distance in average scenarios. Dai and Gluzman~\cite{dai2020queueing,dai2021refined} presented a general bound which depends on the discount factor continuously and studied the application of average-reward RL in a queueing network. Compared with these works, we give a general bound with the discounted and average criteria by using the Kemeny's constant. Besides, we discover the property of the value drifting in average reward value estimation and propose a novel technique called Average Value Constraint to fix it, which we believe is essential to make average DRL algorithms practical in real-world scenarios.

\section{Experiments}
\label{sec:exp}

We conducted a series of experiments to evaluate APO, which are used to answer two questions:
\begin{itemize}
    \item Compared with the discounted PPO, does APO have better performance in environments suitable with average criterion?
    \item What are the factors that affect the performance of APO? Specifically, does APO benefit from the Average Value Constraint?
\end{itemize}
We choose the continuous control benchmark MuJoCo~\cite{mujoco} with the OpenAI Gym ~\cite{brockman2016openai}. The MuJoCo tasks are designed to control the robot to finish certain tasks such as running as fast as possible without falling. It is natural to use the average criterion for some tasks in MuJoCo, such as HalfCheetch in which the robot keeps running and is rewarded by its speed. Moveover, we also evaluate APO in the other experiments with terminal states to examine the generalization ability of APO beyond the theoretical assumptions.

Both APO and PPO are implemented based on a modular RL package in PyTorch named \textit{rlpyt}~\cite{stooke2019rlpyt}. While the performance of PPO is largely dependent on the code-optimizations~\cite{engstrom2020implementation}, we do not consider any tricks apart from the vanilla PPO for controlling variables to justify true impact factors. For each task, we run the algorithm with 5 random seeds for 3 million steps and do the evaluation every 2000 steps. In the evaluation, we run 10 episodes without exploration by setting the standard deviation of policy as zero. All the hyper-parameter combinations we consider are grid searched, which are showed in Appendix~B. The computing infrastructure for running experiments is a server with 2 AMD EPYC 7702 64-Core Processor CPUs and 8 Nvidia GeForce RTX 2080 Ti GPUs. The time consumed of each experiment varies from 1.2h to 2h on average according to the complexity of the task.

\subsection{Comparison to Discounted PPO}
We compare our APO with the standard discounted PPO with the different discounted factors, and select the combination of hyperparameters which achieves the best average performance. The results in Figure~\ref{fig:mujoco} show that APO surpasses all the variants of PPO with different discount factors. We observe that the performances of PPO vary drastically with the change of $\gamma$. Meanwhile, The relationship between performance and $\gamma$ variation is not monotonic. For example, $\gamma=0.99$ achieves a sound result in Ant, while the performance with $\gamma=0.999$ is terrible. This confirms that the performance with the average criterion cannot be optimized by letting $\gamma$ approach 1 in the discounted framework. The experimental results appear to be compatible with the results obtained by Zhang and Ross~\cite{zhang2021average}.

The environments shown in Figure~\ref{fig:mujoco} are suitable for average setting, as they have no terminal states (Swimmer and HalfCheetah) or hard to fall even with a random policy (Ant). It is natural that APO beats the discounted PPO in these tasks. The audience may be curious about the performances of APO in other tasks, such as Hopper, Walker2d and Humanoid. We also evaluate APO in them and show the results in Appendix~C. As we expected, APO achieves sound results with the competent PPO but does not beat the best in Hopper and Walker. However, when we change the metric from average episode return to average reward, we find that APO is better than discounted PPO, especially in Humanoid. It shows that APO achieves higher speed but ignores the terminal reward as it focuses on the average performance under the steady-state distribution. It should be more suitable to model the safety requirements as constraints, which is easy to be extended from APO~\cite{cpo}.

\subsection{Ablation Study on Average Value Constraint}

\begin{figure}[htbp]
    \centering
    \includegraphics[width=\columnwidth]{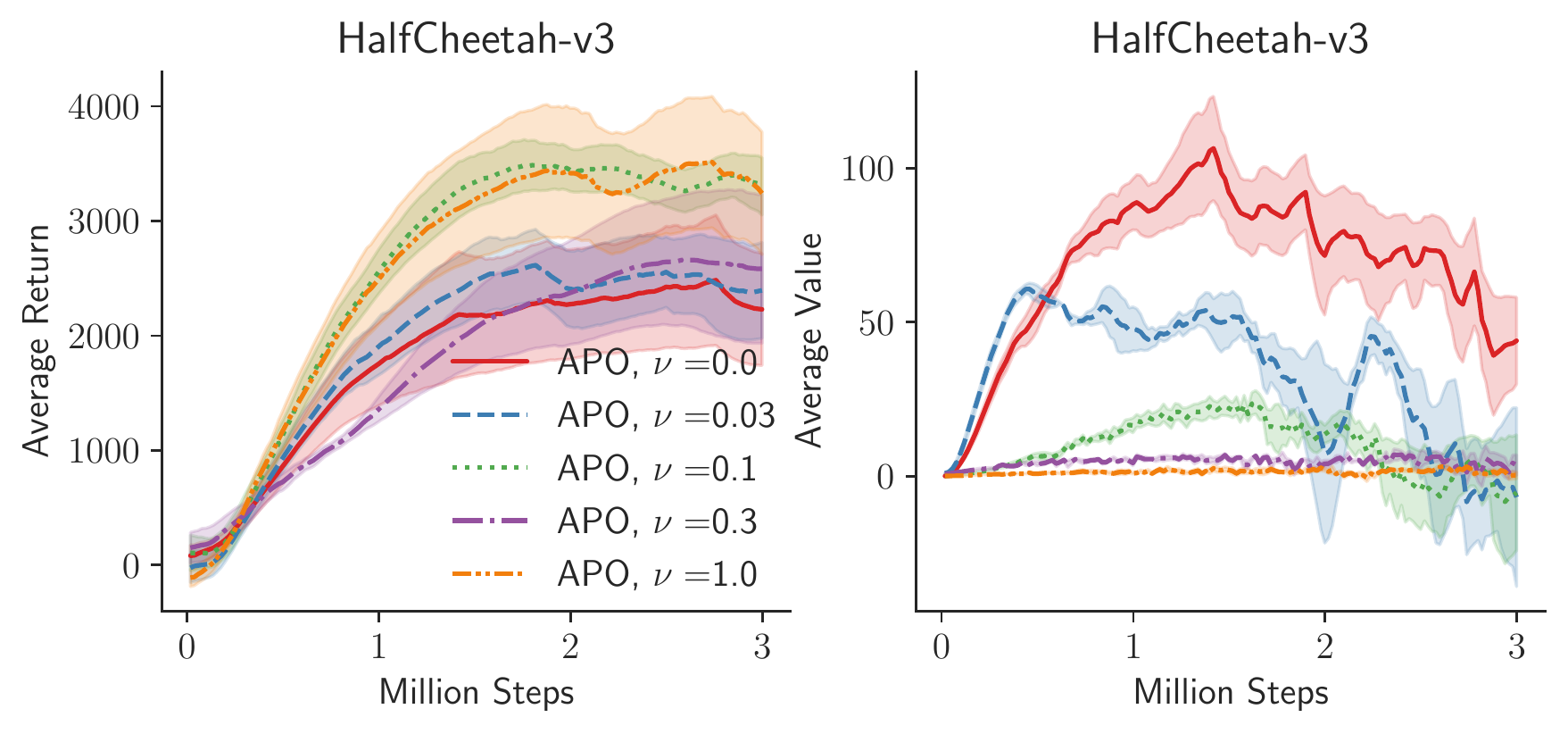}
    \vspace{-2em}
    \caption{Evaluation for different levels $\nu$. On the left and the right are the average return and average value curves respectively.}
    \label{fig:ablation}
\end{figure}

We further evaluate the proposed method Average Value Constraint for the value estimation in average setting. We fix the penalty coefficient $\nu$ at different levels to see the changes of average values during training. The performances are improved by 66.14\% with the Average Value Constraint. As shown in Figure~\ref{fig:ablation}, a larger $\nu$ produces a tighter constraint on the average value. Without the average value constraint ($\nu=0$), the average values fluctuate dramatically during training. Since only the relative value of different states is meaningful for the decision, the bias of average value is detrimental to the policy gradient estimation. As the average value constraint is derived from the Lagrangian method, we can also optimize $\nu$ adaptively during training~\cite{stooke2020responsive}.

\section{Conclusion}
In this paper, we study the trust region approach with the average-reward RL. Based on the PA theory, we prove a lower performance bound with the average criterion and tighten the previous discounted bound with a large discount factor. With the novel lower bound, the monotonic policy improvement of the average trust region method is guaranteed. Furthermore, we analyze the value estimation in the average setting and propose a new technique named Average Value Constraint to stabilize the value estimation during training. In empirical experiments, our proposed algorithm APO achieves better performance than most PPO variants with different discount factors in MuJoCo, demonstrating the average criterion's superiority in many engineering problems. In summary, this work fills the gap in the theory of DRL between the average and discount criteria and provides a practical algorithm with a more appropriate criterion for many practical problems beyond the current benchmarks.

\small
\balance
\bibliographystyle{named}
\bibliography{ijcai21.bib}

\normalsize

\onecolumn
\appendix

\section{Proofs} \label{app:proof}
\subsection{Preliminaries}
\subsubsection{Perturbation Analysis}
\label{sec:app_PA}
Before we present our new result of the performance bound for trust region method, we first review the theory of \textit{Perturbation Analysis}~\cite{cao1997perturbation,cxr}, based on which we give our proofs. Our theory analysis is limited to the finite ergodic Markov chain, which means that from any state it is possible to reach every other state. 

With the discounted reward criterion, we are interested in the discounted reward $\eta_{\pi, \gamma}$, which is defined as
\begin{equation}
    \label{equ:dis_obj}
    \eta_{\pi,\gamma} = (1 - \gamma) \mathbb{E}_{\tau \sim \pi} \left[ \sum_{t=0}^\infty \gamma^t r_t \mid s_0 \sim d_0 \right].
\end{equation}
We further define the \textit{discounted steady-state distribution} of the state $s$ as
\begin{equation}
    \label{equ:dis_dist}
    d_{\pi,\gamma}(s) = (1 - \gamma) \mathbb{E}_{\tau \sim \pi} \left[ \sum_{t=0}^\infty \gamma^t P(s_t = s) \mid s_0 \sim d_0 \right].
\end{equation}
In the matrix form, (\ref{equ:dis_obj}) and (\ref{equ:dis_dist}) are
\begin{align}
    \eta_{\pi, \gamma} &= (1 - \gamma) d_0 \sum_{t=0}^\infty \left(\gamma P_\pi \right)^t r_\pi = (1 - \gamma) d_0 \left( I - \gamma P_\pi \right)^{-1} r_\pi = d_{\pi, \gamma} r_\pi, \\
    d_{\pi,\gamma} &= (1 - \gamma) d_0 \sum_{t=0}^\infty \left(\gamma P_\pi \right)^t = (1 - \gamma) d_0 \left( I - \gamma P_\pi \right)^{-1},
\end{align}
where $d_0$ is the initial state distribution, $P_\pi$ is the transition matrix induced by $\pi$, and $r_\pi$ is the reward vector of $\pi$ with components $r_\pi(s) = \mathbb{E}_{a \sim \pi, s^\prime \sim P_\pi} r(s,a, s^\prime)$.
The core concept of perturbation analysis is the \textit{performance difference formula}:
\begin{equation}
    \label{equ:performance_diff}
    \eta_{\pi^\prime, \gamma} - \eta_{\pi, \gamma} = d_{\pi^\prime, \gamma} \left[ r_{\pi^\prime} - r_\pi + \gamma \left(P_{\pi^\prime} - P_\pi \right) V_{\pi, \gamma} \right],
\end{equation}
where $V_{\pi, \gamma}$ is the \textit{discounted value function} for policy $\pi$ (which is called \textit{discounted potential} in the perturbation analysis). It is known that $V_{\pi, \gamma}$ satisfies the \textit{discounted Poisson equation}:
\begin{equation}
    \label{equ:dis_poisson}
    \left( I - \gamma P_\pi + \gamma e d_\pi \right)V_{\pi, \gamma} = r_\pi,
\end{equation}
where $d_\pi$ is the steady-state distribution and  $e$ is the unit vector. In (\ref{equ:dis_poisson}), the inverse of left matrix $\left( I - \gamma P_\pi + \gamma e d_\pi \right)^{-1}$ is called the \textit{discounted fundamental matrix}, which is denoted as $Z_{\pi, \gamma} $ in this paper. There are some important properties of $Z_{\pi, \gamma}$:
\begin{align}
    Z_{\pi, \gamma} e &= e, \label{equ:Z_property_1}\\
    d_\pi Z_{\pi, \gamma} &= d_\pi, \label{equ:Z_property_2} \\
    Z_{\pi, \gamma} (I - \gamma P_\pi) &= I - \gamma e d_\pi. \label{equ:Z_property_3}
\end{align}
Since all the applications of the value function in optimization depend only on the differences of the components of $V_{\pi, \gamma}$, we can replace $V_{\pi, \gamma}$ with $V_{\pi, \gamma} + c e$, where $c$ could be any constant. In this paper, we choose to calculate $V_{\pi, \gamma}$ as
\begin{equation}
    \label{equ:value_def}
   V_{\pi, \gamma}(s) = \mathbb{E}_{\tau \sim \pi} \left[ \sum_{t=0}^\infty \gamma^t (r_t - \eta_\pi) \mid s_0 = s \right],     
\end{equation}
which satisfies $\mathbb{E}_{s \sim d_\pi} \left[V_{\pi, \gamma}(s) \right] = 0$ (we will give out the proof later).

With the power of performance difference formula, the relationship of (\ref{equ:discounted-identity}) is attained at once by observing that 
\begin{align}
\left[ r_{\pi^\prime} - r_\pi + \gamma \left(P_{\pi^\prime} - P_\pi \right) V_{\pi, \gamma} \right](s) &= \mathbb{E}_{a \sim \pi^\prime, s^\prime \sim P} \left[r(s,a, s^\prime) - \eta_\pi + \gamma V_{\pi, \gamma}(s^\prime) \right] - \mathbb{E}_{a \sim \pi, s^\prime \sim P} \left[r(s,a, s^\prime) - \eta_\pi + \gamma V_{\pi, \gamma}(s^\prime) \right] \\
&= \mathbb{E}_{a \sim \pi^\prime, s^\prime \sim P} \left[r(s,a, s^\prime) - \eta_\pi + \gamma V_{\pi, \gamma}(s^\prime) \right] - V_{\pi, \gamma}(s) \\
&= \mathbb{E}_{a \sim \pi^\prime} \left[A^\pi(s,a) \right]. \label{equ:adv_identity}
\end{align}
Moreover, when $\gamma \to 1$, it is easy to get the limitations of above equations with the average reward criterion:
\begin{itemize}
\item Average reward: $\lim_{\gamma \to 1} \eta_{\pi, \gamma} = \eta_\pi = d_\pi r_\pi$.  
\item Steady-state distribution: $\lim_{\gamma \to 1} d_{\pi, \gamma} = d_\pi$.
\item Average value function: $\lim_{\gamma \to 1} V_{\pi, \gamma} = V_\pi$.
\item Fundamental matrix: $\lim_{\gamma \to 1} Z_{\pi, \gamma} = Z_\pi = \left( I - P_\pi + e d_\pi \right)^{-1}$.
\item Performance difference formula: $\eta_{\pi^\prime} - \eta_{\pi} = d_{\pi^\prime} \left[ r_{\pi^\prime} - r_\pi + \left(P_{\pi^\prime} - P_\pi \right) V_\pi \right]$.
\end{itemize}
Thus, powered by the PA, we are able to extend the analysis of trust region with the discounted reward criterion into the average reward criterion.

\subsubsection{The mean first passage matrix and Kemeny's constant}
Next, we review the concepts of the \textit{mean first passage matrix}~\cite{kemeny1960finite}. If an ergodic Markov chain induced by $\pi$ is started in state $s$, the expected number of steps to reach state $s^\prime$ for the first time is called the \textit{mean first passage time} from $s$ to $s^\prime$, denoted by $M_\pi(s, s^\prime)$. The matrix $M_\pi$ satisfies the following equation:
\begin{equation}
    \label{equ:M_bellman}
    M_\pi = P_\pi (M_\pi - D_\pi) + E,
\end{equation}
where $E$ is a square matrix with all elements being 1 and $D_\pi = (M_\pi)_{\text{dg}}$ satisfying that $D_\pi(s, s) = 1 / d_\pi(s)$ and $D_\pi(s, s^\prime) = 0$ for $s \neq s^\prime$. The subscript 'dg' for some square matrix $N$ refers to a diagonal matrix whose elements are the diagonals of $N$. The mean first passage matrix $M_\pi$ is determined from the fundamental matrix $Z_\pi = \left( I - P_\pi + e d_\pi \right)^{-1}$ by
\begin{equation}
    \label{equ:Z_M}
    M_\pi = \left(I - Z_\pi + E (Z_\pi)_{\text{dg}} \right) D_\pi.
\end{equation}
Kemeny~\shortcite{kemeny1960finite} show that the mean time starting from a given state $s$ to the equilibrium distribution is a constant independent of $s$:
\begin{equation}
    \kappa_\pi = \sum_{s^\prime} d_\pi(s^\prime) M_\pi(s, s^\prime),
\end{equation}
where the constant $\kappa_\pi$ is called \textit{Kemeny's constant}. This constant is an invariant of MDP in the planning horizon.

\subsection{Main Results}

\setcounter{lemma}{0}
\setcounter{theorem}{0}
\setcounter{proposition}{0}

\begin{lemmaA}
    \label{lem:1}
    The matrix $I - (1 - \gamma) M_\pi D_\pi^{-1}$ is invertible when $ (1-\gamma) \kappa_\pi < 1$.
\end{lemmaA}
\begin{proof}
First we will show that $A = \frac{1}{\kappa_\pi} M_\pi D_\pi^{-1}$ is a stochastic matrix. By saying a stochastic matrix, we means that all its elements are non-negative and it satisfies $Ae = e$. By the definition of $M_\pi$ and $D_\pi$, we already know $A$ is a non-negative matrix. Substituting (\ref{equ:Z_M}) into $A$, we have
\begin{equation}
   Ae = \frac{1}{\kappa_\pi} M_\pi D_\pi^{-1} e = \frac{1}{\kappa_\pi} \left(I - Z_{\pi} + E (Z_{\pi})_{\text{dg}} \right) e = \frac{1}{\kappa_\pi}(e - e + \kappa_\pi e) = e.
\end{equation}
With the property of stochastic matrix, we know $\rho(A) = 1$, where $\rho$ is the spectral radius of a square matrix. Further, we have 
 \begin{equation}
     \rho \left((1 - \gamma) M_\pi D_\pi^{-1}\right) < \rho \left(\frac{1}{\kappa_\pi} M_\pi D_\pi^{-1} \right) =\rho(A) = 1.
 \end{equation}
 Again, with the property of stochastic matrix, we know that $I - (1 - \gamma) M_\pi D_\pi^{-1}$ is invertible.
\end{proof}

\begin{lemmaA}
    \label{lem:A2}
    If $ (1-\gamma) \kappa_\pi < 1$, $Z_{\pi, \gamma} = \left[I - M_\pi D_\pi^{-1} + E (Z_{\pi, \gamma} )_{\text{dg}} - (1-\gamma) E(Z_{\pi, \gamma} M_\pi)_{\text{dg}} D_\pi^{-1} \right]\left(I - (1 - \gamma) M_\pi D_\pi^{-1} \right)^{-1}.$
\end{lemmaA}
\begin{proof}
Multiply $\gamma$ on both sides of (\ref{equ:M_bellman}) and rearranging items as follows:
\begin{equation}
    \label{equ:lemma-2-1}
    (1 - \gamma P_\pi) (M_\pi - D_\pi) = \gamma E + (1 - \gamma) M_\pi - D_\pi.
\end{equation}
Left-multiplying $Z_{\pi, \gamma}$, we have
\begin{align}
    Z_{\pi, \gamma} (1 - \gamma P_\pi) (M_\pi - D_\pi) &= \gamma Z_{\pi, \gamma} E + (1 - \gamma) Z_{\pi, \gamma} M_\pi - Z_{\pi, \gamma} D_\pi \\
    &= \gamma E + (1 - \gamma) Z_{\pi, \gamma} M_\pi - Z_{\pi, \gamma} D_\pi, \label{equ:lemma-2-2}
\end{align}
where the equality comes from (\ref{equ:Z_property_2}). Substituting (\ref{equ:Z_property_3}) into the left hand of above equation, we have
\begin{align}
    Z_{\pi, \gamma} (1 - \gamma P_\pi) (M_\pi - D_\pi) &= (I - \gamma e d_\pi) (M_\pi - D_\pi) \\
    &= M_\pi - \gamma e d_\pi M_\pi - D_\pi +  \gamma e d_\pi D_\pi  \\
    &=  M_\pi - \gamma e d_\pi M_\pi - D_\pi +  \gamma E.
\end{align}
Thus we obtain
\begin{equation}
 M_\pi - \gamma e d_\pi M_\pi - D_\pi = (1 - \gamma) Z_{\pi, \gamma} M_\pi - Z_{\pi, \gamma} D_\pi. \label{equ:lemma-2-3}
\end{equation}
When $s = s^\prime$, we have the following relationship with the fact $M_\pi(s,s) = D_\pi(s,s)$:
\begin{equation}
-\gamma [d_\pi M_\pi](s) = (1 - \gamma) [Z_{\pi, \gamma} M_\pi](s,s) - Z_{\pi, \gamma}(s, s) /d_\pi(s).
\end{equation}
Rewriting the above equation into matrix form, we get
\begin{equation}
-\gamma e d_\pi M_\pi = (1 - \gamma) E (Z_{\pi, \gamma} M_\pi)_{\text{dg}} - E (Z_{\pi, \gamma})_{\text{dg}} D_\pi. \label{equ:lemma-2-4}
\end{equation}
Substituting (\ref{equ:lemma-2-4}) into (\ref{equ:lemma-2-3}), we obtain
\begin{equation}
 M_\pi - D_\pi + (1 - \gamma) E (Z_{\pi, \gamma} M_\pi)_{\text{dg}} - E (Z_{\pi, \gamma})_{\text{dg}} D_\pi = (1 - \gamma) Z_{\pi, \gamma} M_\pi - Z_{\pi, \gamma} D_\pi.    
\end{equation}
Right-multiplying both sides by $D_\pi^{-1}$ gives us
\begin{equation}
     Z_{\pi, \gamma}\left(I  - (1 - \gamma) M_\pi D_\pi^{-1} \right)= I - M_\pi D_\pi^{-1} + E (Z_{\pi, \gamma})_{\text{dg}}  - (1 - \gamma) E (Z_{\pi, \gamma} M_\pi)_{\text{dg}} D_\pi^{-1}.
\end{equation}
Finally, with the fact that $I  - (1 - \gamma) M_\pi D_\pi^{-1}$ is invertible by Lemma~\ref{lem:1}, we finish the proof.
\end{proof}

\begin{lemmaA}
    \label{lem:A3}
    $d_{\pi^\prime, \gamma} - d_{\pi, \gamma} = \gamma d_{\pi, \gamma}  (P_{\pi^\prime}-  P_{\pi}) Z_{\pi^\prime, \gamma} $.
\end{lemmaA}
\begin{proof}
By the definition of $d_{\pi^\prime, \gamma} , d_{\pi, \gamma} $, we have
\begin{align}
    d_{\pi^\prime, \gamma} - d_{\pi, \gamma} &= (1 - \gamma) d_0 (I - \gamma P_{\pi^\prime})^{-1} - (1 - \gamma) d_0 (I - \gamma P_{\pi})^{-1} \\
    &= (1 - \gamma) d_0  \left[ (I - \gamma P_{\pi^\prime})^{-1} - (I - \gamma P_{\pi})^{-1} \right] \\
    &= (1 - \gamma) d_0  (I - \gamma P_{\pi})^{-1} \left[ (I - \gamma P_{\pi}) - (I - \gamma P_{\pi^\prime}) \right] (I - \gamma P_{\pi^\prime})^{-1}\\
    &= (1 - \gamma) d_0  (I - \gamma P_{\pi})^{-1} \gamma (P_{\pi^\prime}-  P_{\pi} ) (I - \gamma P_{\pi^\prime})^{-1}\\
    &= \gamma d_{\pi, \gamma} (P_{\pi^\prime}-  P_{\pi} ) (I - \gamma P_{\pi^\prime})^{-1}. \label{equ:diff_dis_dist}
\end{align}
With the fact that $P_{\pi^\prime}e=e$, we obtain
\begin{equation*}
    (I - \gamma P_{\pi^\prime})^{-1} e = \sum_t \left( \gamma P_{\pi^\prime} \right)^t e = \sum_t \left( \gamma^t P_{\pi^\prime}^t e \right) = \sum_t ( \gamma^t e) = \frac{1}{1-\gamma} e.
\end{equation*}
Then, by right-multiplying $I - \gamma P_{\pi^\prime} + \gamma e d_{\pi^\prime}$ to $d_{\pi^\prime, \gamma} - d_{\pi, \gamma} $, we have
\begin{align*}
    (d_{\pi^\prime, \gamma} - d_{\pi, \gamma} ) (I - \gamma P_{\pi^\prime} + \gamma e d_{\pi^\prime} ) &= \gamma d_{\pi, \gamma} (P_{\pi^\prime}-  P_{\pi} ) (I - \gamma P_{\pi^\prime})^{-1}(I - \gamma P_{\pi^\prime} + \gamma e d_{\pi^\prime} ) \\
    &= \gamma d_{\pi, \gamma} (P_{\pi^\prime}-  P_{\pi} )\left(I - \frac{\gamma}{1-\gamma} e d_{\pi^\prime} \right) \\
  &= \gamma d_{\pi, \gamma} (P_{\pi^\prime}-  P_{\pi}). 
\end{align*}
Thus, by the definition of $ Z_{\pi^\prime, \gamma} $, we complete the proof:
\begin{equation}
    \label{equ:lemma4}
    d_{\pi^\prime, \gamma} - d_{\pi, \gamma}  = \gamma d_{\pi, \gamma} (P_{\pi^\prime}-  P_{\pi} ) Z_{\pi^\prime, \gamma}.
\end{equation}
\end{proof}

\begin{proposition}
    For any two stochastic policies $\pi, \pi^\prime$, the following bound holds:
    \begin{align}
        \eta_{\pi^\prime, \gamma} - \eta_{\pi, \gamma} \geq L_{\pi, \gamma}(\pi^\prime) -  2 \epsilon_\gamma D_{\text{TV}}(d_{\pi^\prime, \gamma} \parallel d_{\pi, \gamma}), \\
        \eta_{\pi^\prime, \gamma} - \eta_{\pi, \gamma} \leq L_{\pi, \gamma}(\pi^\prime) +  2 \epsilon_\gamma D_{\text{TV}}(d_{\pi^\prime, \gamma} \parallel d_{\pi, \gamma}),
    \end{align}
where $\epsilon_\gamma = \max_s | \mathbb{E}_{a \sim \pi^\prime} [A_{\pi, \gamma}(s,a)] |$.
\end{proposition}

\begin{proof}
Rewrite $L_{\pi, \gamma}(\pi^\prime)$ in the matrix form:
\begin{equation}
    L_{\pi, \gamma}(\pi^\prime) = d_{\pi, \gamma} \left[ r_{\pi^\prime} - r_\pi + \gamma \left(P_{\pi^\prime} - P_\pi \right) V_{\pi, \gamma} \right].
\end{equation}
Subtracting the above equation from (\ref{equ:performance_diff}), we obtain
\begin{align}
    \left(\eta_{\pi^\prime, \gamma} - \eta_{\pi, \gamma}\right) - L_{\pi, \gamma}(\pi^\prime) &= d_{\pi^\prime, \gamma} \left[ r_{\pi^\prime} - r_\pi + \gamma \left(P_{\pi^\prime} - P_\pi \right) V_{\pi, \gamma} \right] - d_{\pi, \gamma} \left[ r_{\pi^\prime} - r_\pi + \gamma \left(P_{\pi^\prime} - P_\pi \right) V_{\pi, \gamma} \right] \\
    &= (d_{\pi^\prime, \gamma} - d_{\pi, \gamma}) \left[ r_{\pi^\prime} - r_\pi + \gamma \left(P_{\pi^\prime} - P_\pi \right) V_{\pi, \gamma} \right]. \\
\end{align}
Hölder’s inequality tells us that
\begin{equation}
    | \left(\eta_{\pi^\prime, \gamma} - \eta_{\pi, \gamma}\right) - L_{\pi, \gamma}(\pi^\prime) | \leq  \| d_{\pi^\prime, \gamma} - d_{\pi, \gamma} \|_1 \|  r_{\pi^\prime} - r_\pi + \gamma \left(P_{\pi^\prime} - P_\pi \right) V_{\pi, \gamma} \|_\infty = 2 \epsilon_\gamma D_{\text{TV}}(d_{\pi^\prime, \gamma} \parallel d_{\pi, \gamma}),
\end{equation}
where $\epsilon_\gamma = \|  r_{\pi^\prime} - r_\pi + \gamma \left(P_{\pi^\prime} - P_\pi \right) V_{\pi, \gamma} \|_\infty = \max_s | \mathbb{E}_{a \sim \pi^\prime} [A_{\pi, \gamma}(s,a)] |$ following by (\ref{equ:adv_identity}).

\end{proof}

\begin{proposition}
    For any two stochastic policies $\pi, \pi^\prime$, the following bound holds:
    \begin{equation}
        D_{\text{TV}}(d_{\pi^\prime, \gamma} \parallel d_{\pi, \gamma}) \leq \xi_\gamma \mathbb{E}_{s \sim d_{\pi, \gamma} } [D_{\text{TV}}(\pi^\prime \parallel \pi)[s]],
    \end{equation}
where $\xi_\gamma = \min \left\{ \frac{\gamma}{1 - \gamma}, \left| \frac{\gamma(\kappa_{\pi^\prime} -1)}{1 - (1-\gamma)\kappa_{\pi^\prime}} \right| \right\} $ and $\kappa_{\pi^\prime}$ is the Kemeny’s constant of the Markov chain induced by $\pi^\prime$.
\end{proposition}


\begin{proof}
The total variation divergence between $d_{\pi^\prime, \gamma}, d_{\pi, \gamma}$ is alignend with 1-norm of their difference:
\begin{equation}
    D_{\text{TV}}(d_{\pi^\prime, \gamma} \parallel d_{\pi, \gamma}) = \frac{1}{2} \sum_s | d_{\pi^\prime, \gamma}(s) - d_{\pi, \gamma}(s) | = \frac{1}{2} \| d_{\pi^\prime, \gamma} - d_{\pi, \gamma} \|_1.
\end{equation}
Assume that $\frac{\kappa_{\pi^\prime} -1}{\kappa_{\pi^\prime}} < \gamma \leq 1$.
Putting the results of Lemma~\ref{lem:A2} and Lemma~\ref{lem:A3} together leads to
\begin{align}
    d_{\pi^\prime, \gamma} - d_{\pi, \gamma} 
    &= \gamma d_{\pi, \gamma} (P_{\pi^\prime}-  P_{\pi} ) Z_{\pi^\prime, \gamma} \\
    &= \gamma d_{\pi, \gamma} (P_{\pi^\prime}-  P_{\pi} ) \left[I - M_{\pi^\prime} D_{\pi^\prime}^{-1} + E (Z_{\pi^\prime, \gamma} )_{\text{dg}} - (1-\gamma) E(Z_{\pi^\prime, \gamma} M_{\pi^\prime})_{\text{dg}} D_{\pi^\prime}^{-1} \right]\left(I - (1 - \gamma) M_{\pi^\prime} D_{\pi^\prime}^{-1} \right)^{-1} \\
    &= \gamma d_{\pi, \gamma} (P_{\pi^\prime}-  P_{\pi} ) \left( I - M_{\pi^\prime} D_{\pi^\prime}^{-1} \right) \left(I - (1 - \gamma) M_{\pi^\prime} D_{\pi^\prime}^{-1} \right)^{-1},
\end{align}
where the last equality comes from the fact that $P_{\pi^\prime} E =  P_{\pi} E = E$. Comparing the previous result (see Lemma 3 in~\cite{zhang2021policy}), we obtain an extra term $\gamma \left(I - (1 - \gamma) M_{\pi^\prime} D_{\pi^\prime}^{-1} \right)^{-1}$, which is negligible when $\gamma=1$ but meaningful with large discount factors.

Thus, we have 
\begin{align}
    \| d_{\pi^\prime, \gamma} - d_{\pi, \gamma} \|_1 
    &= \| \gamma d_{\pi, \gamma} (P_{\pi^\prime}-  P_{\pi} ) \left( I - M_{\pi^\prime} D_{\pi^\prime}^{-1} \right) \left(I - (1 - \gamma) M_{\pi^\prime} D_{\pi^\prime}^{-1} \right)^{-1} \|_1 \\
    &\leq \gamma \| d_{\pi, \gamma} (P_{\pi^\prime}-  P_{\pi} ) \|_1 \| \left( I - M_{\pi^\prime} D_{\pi^\prime}^{-1} \right) \left(I - (1 - \gamma) M_{\pi^\prime} D_{\pi^\prime}^{-1} \right)^{-1} \|_\infty \\
    &\leq \gamma \| d_{\pi, \gamma} (P_{\pi^\prime}-  P_{\pi} ) \|_1 \|  \left( I - M_{\pi^\prime} D_{\pi^\prime}^{-1} \right)\|_\infty \| \left(I - (1 - \gamma) M_{\pi^\prime} D_{\pi^\prime}^{-1} \right)^{-1} \|_\infty,
\end{align}
where the first inequality follows from the Hölder’s inequality. 
Next, let us bound the items one by one.
\begin{align}
\| d_{\pi, \gamma} (P_{\pi^\prime}-  P_{\pi} ) \|_1 &=\sum_{s^{\prime}}\left|\sum_{s}\left(\sum_{a} (P\left(s^{\prime} \mid s, a\right) \pi^{\prime}(a \mid s)-P\left(s^{\prime} \mid s, a\right) \pi(a \mid s) ) \right) d_{\pi, \gamma}(s)\right| \\
& \leq \sum_{s^{\prime}, s}\left|\sum_{a} P\left(s^{\prime} \mid s, a\right)\left(\pi^{\prime}(a \mid s)-\pi(a \mid s)\right)\right| d_{\pi, \gamma}(s) \\
& \leq \sum_{s, s^{\prime}, a} P\left(s^{\prime} \mid s, a\right)\left|\pi^{\prime}(a \mid s)-\pi(a \mid s)\right| d_{\pi, \gamma}(s) \\
& \leq \sum_{s, a}\left|\pi^{\prime}(a \mid s)-\pi(a \mid s)\right| d_{\pi, \gamma}(s) \\
&=2 \underset{s \sim d_{\pi, \gamma}}{\mathbb{E}}\left[D_{\mathrm{TV}}\left(\pi^{\prime} \parallel \pi\right)[s] \right].
\end{align}

\begin{equation}
\left\| \left( I - M_{\pi^\prime} D_{\pi^\prime}^{-1} \right) \right\|_\infty = \sum_s \left[ \sum_{s^\prime} M(s, s^\prime) d_{\pi^\prime}(s) - 1\right] = \kappa_{\pi^\prime} - 1.    
\end{equation}

\begin{align}
\left\| \left(I - (1 - \gamma) M_{\pi^\prime} D_{\pi^\prime}^{-1} \right)^{-1} \right\|_\infty 
&= \left\| \sum_{t=0}^\infty \left( (1 - \gamma) M_{\pi^\prime} D_{\pi^\prime}^{-1} \right)^t \right\|_\infty \\
&\leq \sum_{t=0}^\infty \left\| (1 - \gamma) M_{\pi^\prime} D_{\pi^\prime}^{-1} \right\|_\infty^t \\
&= \sum_{t=0}^\infty \left( (1 - \gamma) \kappa_{\pi^\prime} \right)^t \\
&= \frac{1}{(1 - \gamma) \kappa_{\pi^\prime}}.
\end{align}

Combining the bounds above, we conclude that:
\begin{align}
    \| d_{\pi^\prime, \gamma} - d_{\pi, \gamma} \|_1 
    &= \| \gamma d_{\pi, \gamma} (P_{\pi^\prime}-  P_{\pi} ) \left( I - M_{\pi^\prime} D_{\pi^\prime}^{-1} \right) \left(I - (1 - \gamma) M_{\pi^\prime} D_{\pi^\prime}^{-1} \right)^{-1} \|_1 \\
    &\leq \gamma \| d_{\pi, \gamma} (P_{\pi^\prime}-  P_{\pi} ) \|_1 \| \left( I - M_{\pi^\prime} D_{\pi^\prime}^{-1} \right) \left(I - (1 - \gamma) M_{\pi^\prime} D_{\pi^\prime}^{-1} \right)^{-1} \|_\infty \\
    &\leq \gamma \| d_{\pi, \gamma} (P_{\pi^\prime}-  P_{\pi} ) \|_1 \|  \left( I - M_{\pi^\prime} D_{\pi^\prime}^{-1} \right)\|_\infty \| \left(I - (1 - \gamma) M_{\pi^\prime} D_{\pi^\prime}^{-1} \right)^{-1} \|_\infty \\
     &\leq \frac{\gamma(\kappa_{\pi^\prime} -1)}{1 - (1-\gamma)\kappa_{\pi^\prime}} \mathbb{E}_{s \sim d_{\pi, \gamma} } [D_{\text{TV}}(\pi^\prime \parallel \pi)[s]]. \label{equ:bound_right}
\end{align}
Similar with the above analysis, when $\gamma < 1$, we conclude the other part of the bound from (\ref{equ:diff_dis_dist})
\begin{align}
    \| d_{\pi^\prime, \gamma} - d_{\pi, \gamma} \|_1 
    &= \| \gamma d_{\pi, \gamma} (P_{\pi^\prime}-  P_{\pi} ) (I - \gamma P_{\pi^\prime})^{-1} \|_1 \\
    &\leq \gamma \| d_{\pi, \gamma} (P_{\pi^\prime}-  P_{\pi} ) \|_1 \| (I - \gamma P_{\pi^\prime})^{-1} \|_\infty \\
     &\leq \frac{\gamma}{1-\gamma} \mathbb{E}_{s \sim d_{\pi, \gamma} } [D_{\text{TV}}(\pi^\prime \parallel \pi)[s]],
\end{align}
where
\begin{equation}
\left\| (I - \gamma P_{\pi^\prime})^{-1} \right\|_\infty
= \left\| \sum_{t=0}^\infty \left(  \gamma P_{\pi^\prime} \right)^t \right\|_\infty \leq \sum_{t=0}^\infty \gamma^t \left\| P_{\pi^\prime} \right\|_\infty^t \leq \sum_{t=0}^\infty \gamma^t  =\frac{1}{1 - \gamma}. \label{equ:bound_left}
\end{equation}
Thus we recover the bound in~\cite{cpo}.
Note that (\ref{equ:bound_left}) is only meaningful for $\gamma < 1$ and (\ref{equ:bound_right}) is only meaningful for $\frac{\kappa_{\pi^\prime} -1}{\kappa_{\pi^\prime}} < \gamma \leq 1$.
When $\gamma < \frac{\kappa_{\pi^\prime} -1}{\kappa_{\pi^\prime}}$, we have
\begin{equation}
    \left| \frac{\gamma(\kappa_{\pi^\prime} -1)}{1 - (1-\gamma)\kappa_{\pi^\prime}} \right| =  \frac{\gamma(\kappa_{\pi^\prime} -1)}{(1-\gamma)\kappa_{\pi^\prime} - 1} = \frac{\gamma}{1 - \gamma} \left(\frac{\kappa_{\pi^\prime} -1}{\kappa_{\pi^\prime} - 1 / (1-\gamma)}  \right) > \frac{\gamma}{1 - \gamma}.
\end{equation}
Thus we combine the bounds with $\xi_\gamma = \min \left\{ \frac{\gamma}{1 - \gamma}, \left| \frac{\gamma(\kappa_{\pi^\prime} -1)}{1 - (1-\gamma)\kappa_{\pi^\prime}} \right| \right\}$ in a unified form.
\end{proof}


\begin{theorem}
    For any two stochastic policies $\pi, \pi^\prime$, the following bound holds:
    \begin{align*}
        \eta_{\pi^\prime, \gamma} - \eta_{\pi, \gamma} \geq L_{\pi, \gamma}(\pi^\prime) -  2 \epsilon_\gamma \xi_\gamma \mathbb{E}_{s \sim d_{\pi, \gamma} } [D_{\text{TV}}(\pi^\prime \parallel \pi)[s]], \\
        \eta_{\pi^\prime, \gamma} - \eta_{\pi, \gamma} \leq L_{\pi, \gamma}(\pi^\prime) +  2 \epsilon_\gamma \xi_\gamma \mathbb{E}_{s \sim d_{\pi, \gamma} } [D_{\text{TV}}(\pi^\prime \parallel \pi)[s]].
    \end{align*}
    In particular, the bounds hold with the average criterion:
    \begin{align*}
        \eta_{\pi^\prime} - \eta_\pi \geq L_\pi(\pi^\prime) - 2 \epsilon \xi \mathbb{E}_{s \sim d_\pi } [D_{\text{TV}}(\pi^\prime \parallel \pi)[s]], \\
        \eta_{\pi^\prime} - \eta_\pi \leq L_\pi(\pi^\prime) + 2 \epsilon \xi \mathbb{E}_{s \sim d_\pi } [D_{\text{TV}}(\pi^\prime \parallel \pi)[s]].
    \end{align*}
\end{theorem}
\begin{proof}
The theorem is the combination of two above propositions.
\end{proof}

\begin{proposition}
    $\mathbb{E}_{s \sim d_\pi} \left[V_{\pi, \gamma}(s) \right] = 0$.    
\end{proposition}
\begin{proof}
Revisit the definition of $V_{\pi, \gamma}$ in matrix form,
\begin{equation}
    V_{\pi, \gamma} = \sum_{t=0}^\infty \left( \gamma P_\pi \right)^t (r_\pi - e \eta_\pi) = r_\pi - e d_\pi r_\pi + \sum_{t=1}^\infty \left( \gamma^t (P_\pi  - e d_\pi)^t r_\pi \right)  = \left[(I - \gamma P_\pi + \gamma e d_\pi)^{-1} - e d_\pi \right]r_\pi.
\end{equation}
With the fact that $d_\pi P_\pi = d_\pi$ and $d_\pi e = 1$, we have
\begin{align}
    \mathbb{E}_{s \sim d_\pi} V_{\pi, \gamma}(s) &= d_\pi V_{\pi, \gamma} \\
    &= d_\pi \left[ (I - \gamma P_\pi + \gamma e d_\pi)^{-1} - e d_\pi \right] r_\pi \\
    &=  d_\pi \left[I + \sum_{t=1}^{\infty} \left( \gamma^t  (P_\pi - e d_\pi)^t \right) - e d_\pi \right] r_\pi \\
    &=  d_\pi \left[I + \sum_{t=1}^{\infty} \left( \gamma^t  (P_\pi^t - e d_\pi) \right) - e d_\pi \right] r_\pi \\
    &=  \left[d_\pi + \sum_{t=1}^{\infty} \left( \gamma^t  (d_\pi - d_\pi) \right) -  d_\pi \right] r_\pi \\
    &= 0.
\end{align}
\end{proof}

\section{Hyperparameters of APO and PPO}
\label{sec:hyperparameter}

\begin{table}[H]
\centering
\begin{tabular}{lr}
\toprule
Hyperparameter & Value \\
\midrule
\emph{Shared} \\
\hspace{1em} Network learning rate $\beta$ & 3e-4 \\
\hspace{1em} Network hidden sizes  & [64, 64] \\
\hspace{1em} Activation function & Tanh \\
\hspace{1em} Optimizer & Adam \\
\hspace{1em} Batch size & 256 \\
\hspace{1em} Gradient Clipping & 10 \\
\hspace{1em} Clipping parameter $\varepsilon$ & 0.2 \\
\hspace{1em} Optimization Epochs $M$ & 10 \\
\hspace{1em} GAE parameter $\lambda$ & [0.8, 0.9, 0.95, 0.99] \\
\midrule
\emph{PPO} \\
\hspace{1em} Discount factor $\gamma$ & [0.9, 0.95, 0.99, 0.999] \\
\midrule
\emph{APO} \\
\hspace{1em} Step size $\alpha$ & [0.03, 0.1, 0.3] \\
\hspace{1em} Average Value Constraint Coefficient $\nu$ & [0, 0.03, 0.1, 0.3, 1.0] \\
\bottomrule
\end{tabular}
\caption{Hyperparameters Sheet}
\label{tab:hyper}
\end{table}

\section{Additional Results} 
\begin{figure*}[htbp]
    \centering
    \includegraphics[width=\linewidth]{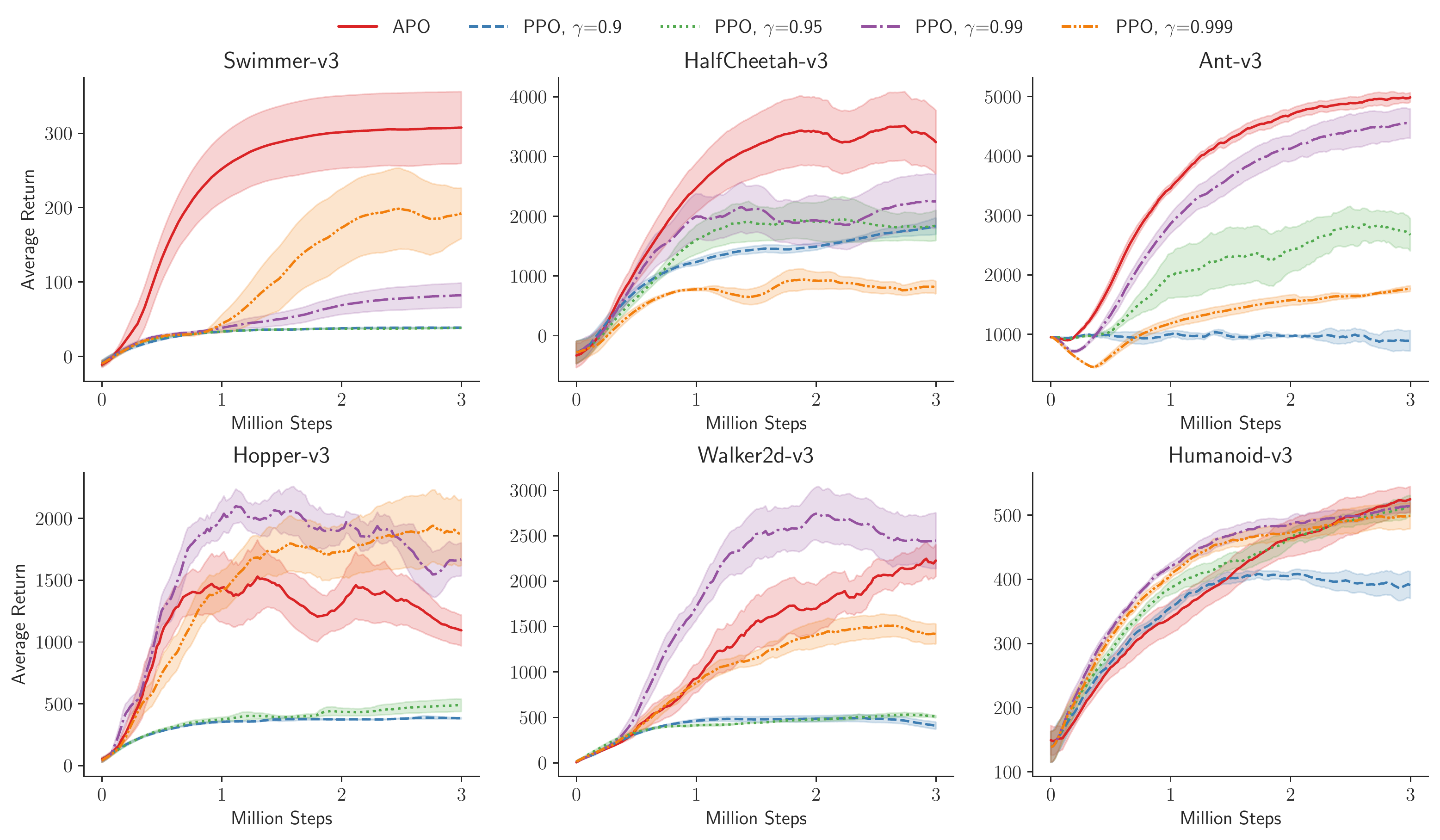}
    \caption{Additional comparison of APO and PPO for the average return in MuJoCo tasks.}
    \label{fig:mujoco_return}
\end{figure*}

\begin{figure*}[htbp]
    \centering
    \includegraphics[width=\linewidth]{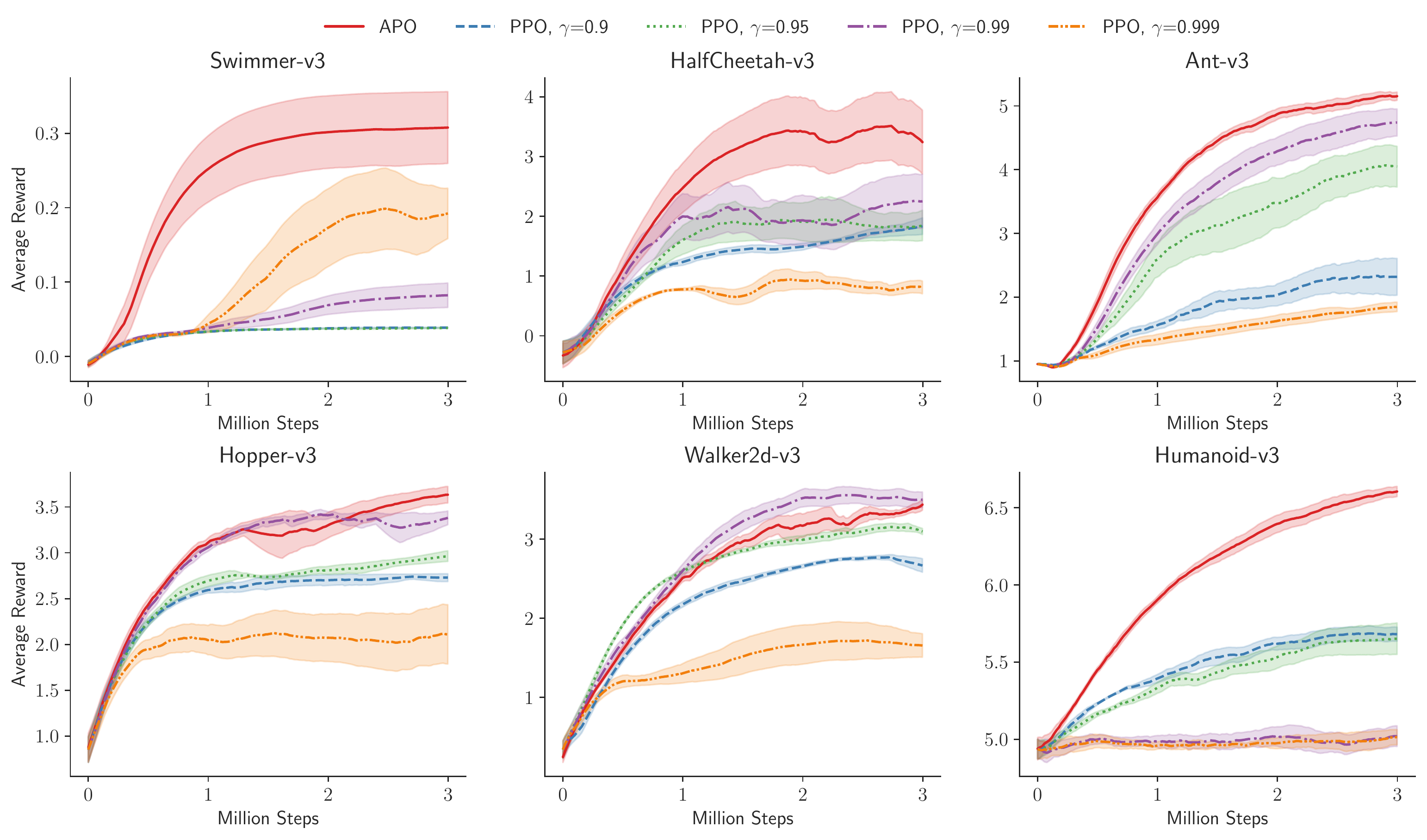}
    \caption{Additional comparison of APO and PPO for the average reward in MuJoCo tasks.}
    \label{fig:mujoco_reward}
\end{figure*}

We benchmark our method on the MuJoCo continuous control tasks. The training curves of average episode return and average reward are shown in Figure~\ref{fig:mujoco_return} and Figure~\ref{fig:mujoco_reward} respectively. The environments Swimmer and HalfCheetah do not have terminal states, so the treads of curves are the same in the two metrics. In these tasks, APO has the best performance in both the episode return and average reward. 
A similar result also exists in Ant, in which the multi-legged robot is safe from falling. In the other tasks with the unsafe states, APO still has better performances in terms of average reward than discounted PPO, especially in Humanoid.
There is no evidence that the discounted criterion is better than the average one in problems with unsafe states though APO does not beat the best PPO in terms of average episode return in Hopper and Walker. As we mentioned in Section~\ref{sec:exp}, the safety problem in MuJoCo is more suitable to be formulated as a constrained problem optimizing the average reward, which is a promising direction for future research.
\label{sec:additional_results}

\end{document}